\documentclass{article}
\usepackage{arxiv}

\usepackage{tabularx}
\usepackage{multirow}
\usepackage{array}

\usepackage[dvipsnames]{xcolor}
\usepackage{censor}
\usepackage[utf8]{inputenc} 
\usepackage[T1]{fontenc}    
\usepackage{hyperref} 

\definecolor{note}{HTML}{235c1f}
\usepackage{graphicx}
\usepackage[export]{adjustbox}

\usepackage{url}            
\usepackage{booktabs}       

\usepackage{lipsum}	
\usepackage{pifont}

\usepackage{mathtools}

\usepackage{multirow}
\usepackage{multicol}
\usepackage{paralist}
\usepackage{float}
\usepackage[numbers]{natbib}

\title{Validity of Feature Importance in Low-Performing Machine Learning for Tabular Biomedical Data}
\author{
Youngro Lee\\
  \small{Seoul National University}\\
  \texttt{youngro12@snu.ac.kr} \\
  \And
  Giacomo Baruzzo\\
  \small{University of Padova}\\
  \And
  Jeonghwan Kim\\
  \small{Georgia Institute of Technology}\\
\And
Jongmo Seo \\
  \small{Seoul National University}\\
\And
Barbara Di Camillo \\
  \small{University of Padova}\\
}

\begin{document}
\maketitle

\begin{abstract}
In tabular biomedical data analysis, tuning models to high accuracy is considered a prerequisite for discussing feature importance, as medical practitioners expect the validity of feature importance to correlate with performance. In this work, we challenge the prevailing belief, showing that low-performing models may also be used for feature importance. We propose experiments to observe changes in feature rank as performance degrades sequentially. Using three synthetic datasets and six real biomedical datasets, we compare the rank of features from full datasets to those with reduced sample sizes (data cutting) or fewer features (feature cutting). In synthetic datasets, feature cutting does not change feature rank, while data cutting shows higher discrepancies with lower performance. In real datasets, feature cutting shows similar or smaller changes than data cutting, though some datasets exhibit the opposite. When feature interactions are controlled by removing correlations, feature cutting consistently shows better stability. By analyzing the distribution of feature importance values and theoretically examining the probability that the model cannot distinguish feature importance between features, we reveal that models can still distinguish feature importance despite performance degradation through feature cutting, but not through data cutting. We conclude that the validity of feature importance can be maintained even at low performance levels if the data size is adequate, which is a significant factor contributing to suboptimal performance in tabular medical data analysis. This paper demonstrates the potential for utilizing feature importance analysis alongside statistical analysis to compare features relatively, even when classifier performance is not satisfactory.

\end{abstract}

\section{Introduction}
\subsection{Background}

Exploring feature importance is crucial in biomedical data analysis. Recently, machine learning methods have outperformed conventional regression across various domains. Consequently, identifying important features through machine learning model interpretation is increasingly applied in research and industry to understand each feature's impact on the model outcome, aiding interpretability. This process is applicable across fields but is especially critical in biomedical applications, where feature importance can be the primary goal. In bioinformatics, it helps identify biomarkers among extensive genetic data points \citep{wang2021characteristics, thomas2019metagenomic, lee2023machine, aryal2020machine}. In medical fields, important features guide the understanding of disease symptoms or causes and can validate decision support systems before clinical use \citep{hung2023developing, tso2022machine, zhang2020machine}.

Behind this popularity, the methodological approach heavily depends on the type of dataset used. For predictive models, various deep neural networks specialized for specific tasks are studied across multi-dimensional datasets, such as images, time-series data, or text. In neural network-based methods, there is no clear standard for interpretation, leading to the proposal of numerous methodologies and ongoing active discussions in the field of machine learning \citep{selvaraju2020grad, chefer2021transformer, hooker2019benchmark, covert2020understanding}. However, tabular datasets typically consist of much lower dimensions. Even in cases of high-dimensional data, the data is often compressed into lower dimensions through various feature selection methods \citep{lee2023machine,khaire2022stability}. Consequently, compared to other types of datasets, typical machine learning models and interpretation methods are more commonly used rather than developing novel structures. In the biomedical field, tree-ensemble models, such as Random Forest, XGBoost (Extreme Gradient Boosting), and LGBM (Light Gradient Boosting Model), have gained popularity due to their high performance and less computation burden compared to neural network. To interpret these models, intrinsic methods like Gini impurity or model-agnostic methods like SHAP (Shapley Additive exPlanations) are commonly employed \citep{lundberg2017unified, lee2024cleshcomprehensiveliteralexplanation}.

\subsection{Related Works}
Unlike statistics, feature importance from machine learning typically lacks validation steps\cite{lee2023machine}. Discussions on factors affecting the validity of feature importance highlight issues such as high correlations between features, which can mislead interpretations \cite{tolocsi2011classification, hooker2021unrestricted}, and biases arising from feature properties (categorical or continuous) and the number of categories \cite{strobl2007bias}. Lower performance tends to flatten the distribution of feature importance due to these biases\cite{leino2018feature}.

Despite the many opportunities for discussion regarding the validity of feature importance, the current application of machine learning and feature importance for tabular datasets in biomedical fields often aligns the ranking of feature importance without thoroughly analyzing its validity \citep{lee2023suggestion, lee2024cleshcomprehensiveliteralexplanation}. Instead of critically evaluating the validity of feature importance, high model accuracy is generally considered a prerequisite for discussing feature importance, with the assumption that its validity is correlated with model performance. This belief tends to generalize the methodological pipeline, which can hinder further discussion and limit the use of machine learning in cases of weaker performance, even though statistical analysis is still conducted.

However, no direct experimental studies have explored this relationship. Why has this relationship not been clarified? Tracking feature importance relative to classification performance requires a measure to define feature importance error. However, the correct order of feature importance is subjective and possibly nonexistent. Validity may be measured indirectly: when noise features are independent of a label, the ability to filter out these noise features without confusing them with important features indicates validity \cite{janitza2013auc}. However, to observe validity patterns regarding performance degradation, the overall feature importance validity must be calculated numerically and comparably. An intuitive validity index, similar to accuracy, would aid public understanding.

\subsection{Proposed analysis framework}
We investigate how the validity of feature importance varies with the performance of machine learning on tabular datasets. We investigate by generating three synthetic datasets with varying label balances and collecting six biomedical datasets of varying sample size, feature number and sources. As tabular datasets are represented as 2-dimensional matrices with samples and features, performance issues can arise either from insufficient data size or inadequate number of features. 

Using Random Forest as a prediction model, classification accuracy is measured using the area under the ROC curve (AUC). We compare results from full datasets to those with reduced sample sizes (data cutting) or fewer features (feature cutting) using controlled degradation algorithms. To assess feature selection stability,  we use  stability indexes,rank difference, Spearman’s rank correlation (SRCC), Canberra distance (CD), and Bray–Curtis distance \cite{lee2023machine, khaire2022stability}. Additional experiments investigate the role of correlated features and feature importance variability. A theoretical analysis shows that, under specific hypothesis, machine learning models tend to misinterpret feature ranks due to insufficient data size more than lack of features.

In conclusion, we suggest that the validity of feature importance can be maintained even at low performance levels if the low performance is due to a lack of features, but not a lack of samples.

\section{Data and Methods}
\subsection{Data}

\begin{table}[h!]
\centering
\small 
\setlength{\tabcolsep}{4pt} 
\renewcommand{\arraystretch}{1.2} 
\caption{\textbf{Characteristics of datasets}}
\resizebox{\textwidth}{!}{
\begin{tabular}{|p{3cm}|p{2cm}|p{2cm}|p{2cm}|p{2cm}|p{2cm}|p{2cm}|p{2cm}|p{2cm}|p{2cm}|}
\hline
 & \multicolumn{3}{|c|}{Synthetic Dataset} & \multicolumn{6}{|c|}{Real Dataset} \\
\hline
 & Generated Dataset 1 & Generated Dataset 2 & Generated Dataset 3 & Real Dataset 1 & Real Dataset 2 & Real Dataset 3 & Real Dataset 4 & Real Dataset 5 & Real Dataset 6 \\
\hline
Reference &  &  &  & \cite{lee2023machine} & \cite{antal2014diabetic} & \cite{https://doi.org/10.24432/c5z89r} & \cite{https://doi.org/10.24432/c5z60n}& \cite{blake1998uci} & \cite{patel2015reliable} \\
\hline
Target & \multicolumn{3}{|c|}{Linear combination $>$ standard?} & IBD & Diabetic Retinopathy & Heart Failure & Survival after Thoracic Surgery & Heart Abnormality & Breast Cancer \\
\hline
\# of samples & 10000 & 10000 & 10000 & 1569 & 1151 & 299 & 470 & 267 & 198 \\
\hline
\# of positive samples (\%) & 5085 (50.8) & 2501 (25.0) & 7501 (75.0) & 702 (44.7) & 611 (53.1) & 96 (32.1) & 70 (14.9) & 212 (79.4) & 47 (23.7) \\
\hline
\# of negative samples (\%) & 4915 (49.2) & 7499 (75.0) & 2499 (25.0) & 867 (55.3) & 540 (46.9) & 203 (67.9) & 400 (85.1) & 55 (20.6) & 151 (76.3) \\
\hline
\end{tabular}
}
\label{table:1}
\end{table}

\textit{\textbf{Synthetic Datasets}} 

To create a dataset with clear feature rankings, we generate a simulated dataset with 
 $N=20$ features and 10,000 samples each, where binary labels $\{0, 1\}$ are obtained based on a linear combination of independent features. Specifically, we obtain each feature sample 
 $X = [x_1, x_2, \ldots, x_N]$, with $0\leq x_n \leq 1$ for $n=1,\ldots, N$. Let $A=[a_1, a_2, \ldots, a_N]$ be the coefficient vector, where $a_n \propto n$. Then we define $y = \sum_{n=1}^{N}a_n x_n = A \cdot X$.

When \(y\) passes a pre-defined threshold, the sample is assigned class label 1; otherwise, 0 (see Appendix 1). For the simulated dataset, the feature ranking is obvious because each feature is independent but has a stationary coefficient value that would be proportionate to feature importance.

\textit{\textbf{Real Datasets}} 

Real datasets have much higher complexities than that of a simulation dataset. Features have complicated interactions, feature values are not equally distributed and feature importance is not easily definable. In Table 1, six different types of real medical datasets, which are all open-sourced, are presented; they have different number of features, dataset size, and characteristics. Description of each dataset is available in Appendix 2.

\subsection{Methods}
\begin{figure*}
    \centering
    \includegraphics[width=0.95\textwidth]{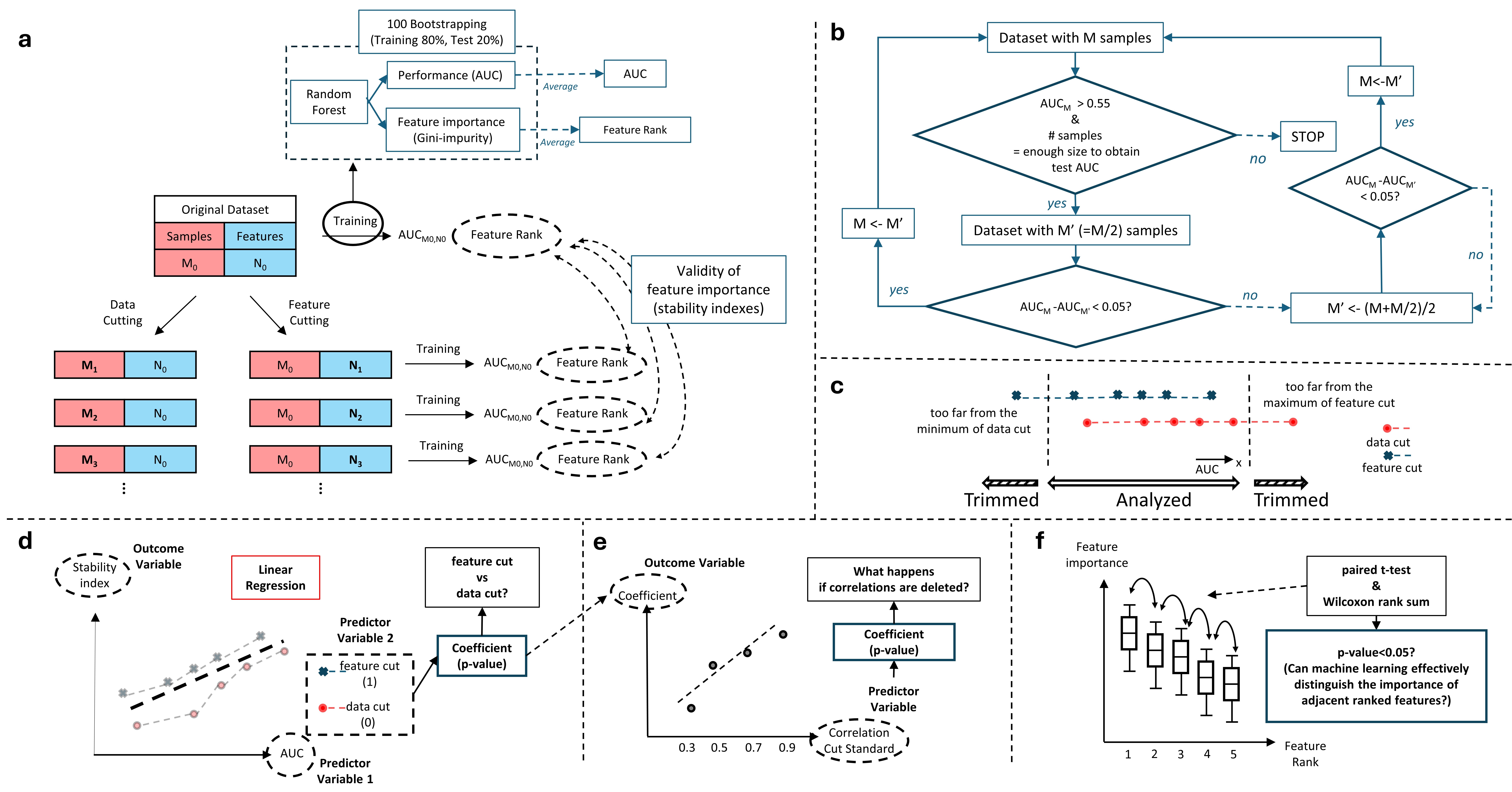}
    \caption{\textbf{Graphical explanation for overall experiments} a) Overall structure how the performance and stability indexes are obtained. b) Algorithm for data cutting c) Schematic for trimming results, d) Schematic of linear regression to compare two degradation algorithms (Section 3.1) e) Schematic of linear regression to analyze the effect of deleting correlations (Section 3.2) f) Schematic of feature importance distribution analysis (Section 3.3) }
\end{figure*}

The overall design of experiments is illustrated in Figure 1. In Figure 1a, the experiment begins by training the original dataset using a Random Forest and calculating the AUC and feature rank by averaging the results across 100 bootstraps. The number of samples or features is then reduced using data cutting and feature cutting algorithms, as shown in Figure 1b. Each reduced dataset is trained the same way, and the resulting feature ranks are compared to those from the original dataset to calculate stability indexes; namely: rank difference, Spearman’s rank correlation, Canberra distance, and Bray–Curtis distance. Details on the use of Random Forest (how it is applied and why Random Forest among many machine learning algorithms) are available in Appendix 3 \cite{nembrini2018revival}
. The performance index (AUC) is described in Appendix 4 \cite{hosmer2000applied, mandrekar2010receiver, patel2015reliable}, and the stability index algorithm, including rank difference calculation, in Appendix 5 \cite{lee2023machine, khaire2022stability}. Used Python library packages and devices are listed in Appendix 6.

The data and feature cutting algorithms are described in Figure 1b. The algorithm starts with the full dataset, conducting experiments with half the current sample size, then setting this as the next sample size. If the AUC difference between two consecutive iterations exceed 0.05, another experiment is conducted with the average of the current sample size, setting this as the next sample size. This process is repeated  until the AUC drops below 0.55 or the sample size is too small to calculate AUC due to missing classes in the test labels of any bootstrap. The same algorithm is applied for feature cutting, where features are cut sequentially by importance ranks from the initial experiment, starting with the most important.

As illustrated in Figure 1c, the results from the real dataset are trimmed due to the limited number and variability across performance ranges. Specifically, to avoid outliers and ensure fair comparison, no more than one data point from another algorithm is included beyond the maximum and minimum performance data points of each algorithm.

Validity comparison between performance degradation algorithms (Section 3.1) is done by plotting and linear regression as in Figure 2d. Since results from two degradation algorithms are not paired, linear regression is used to assess the effect of the degradation algorithm choice. We also analyze results after sequentially deleting correlated features to reduce feature interaction (Section 3.2). We calculate the degradation algorithm’s coefficient in each dataset where correlations are deleted, and regress the coefficient value by its correlation cut standard (Figure 1e). In Section 3.3, we compare feature importance values between adjacently ranked features in each experiment (Figure 1f).

\section{Results}
\subsection{Validity Comparison Between Data Cutting and Feature Cutting}
 
\subsubsection{Synthetic Datasets}

\begin{figure*}
    \centering
    \includegraphics[width=0.8\textwidth]{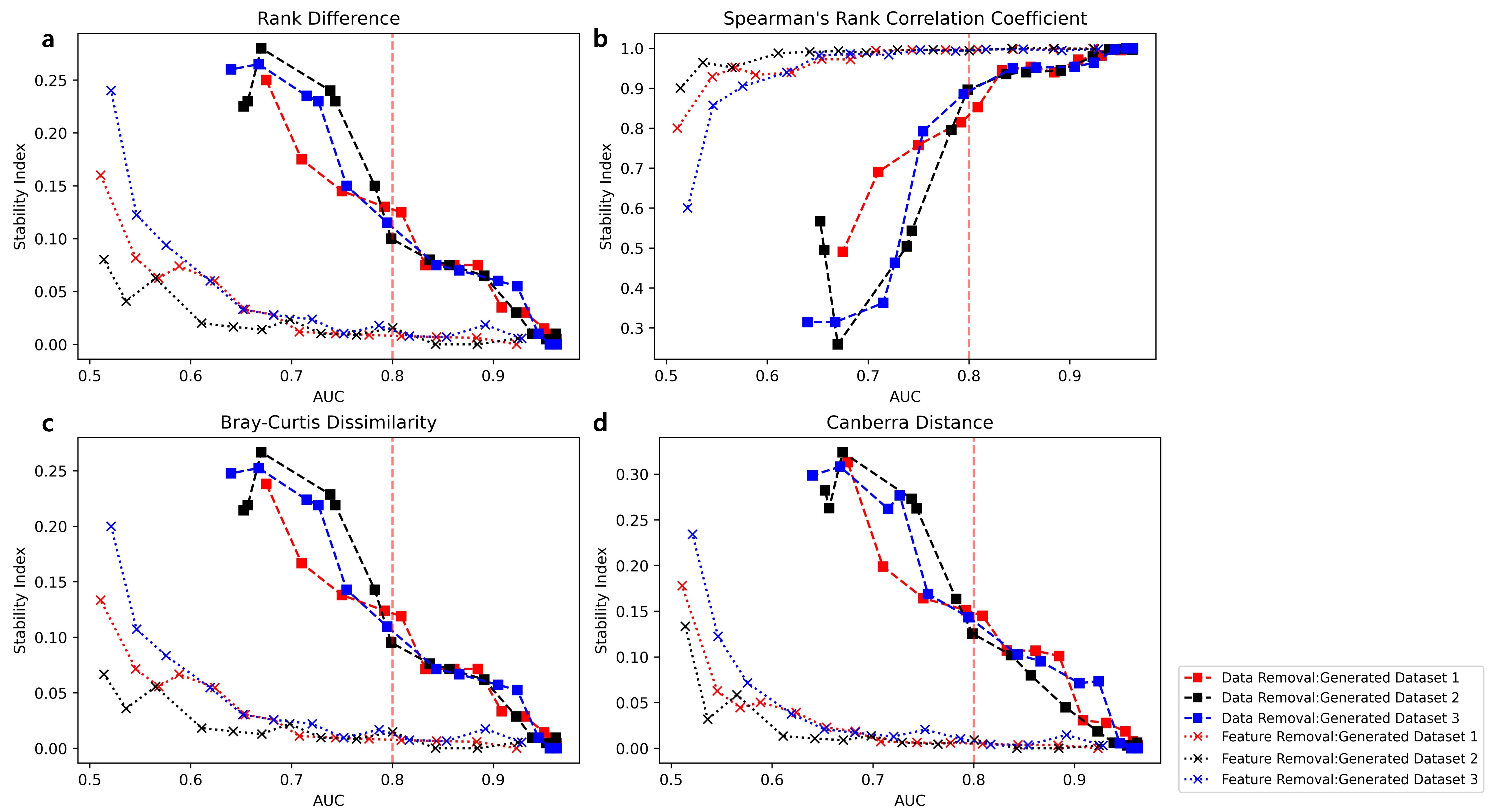}
    \caption{\textbf{Comparison between data cutting and feature cutting regarding four stability indexes in experiment on synthetic datasets.} x-axis refers to AUC value and y axis to stability index. Red vertical line refers to x (AUC) = 0.8 to denote the point where the stability index starts to rapidly change. }
\end{figure*}

The relationship between AUC and stability index in each degradation algorithm shows a consistent pattern regardless of the generated dataset type or stability index. The stability index drops consistently with performance degradation through data cutting. While it diminishes slowly when AUC is, roughly above 0.8, it drops rapidly when AUC falls below 0.8. In contrast, feature cutting shows no clear relationship between performance and stability index. Even with AUC in 0.6 to 0.7, the stability index remains similar to that at higher AUC values. By linear regression in Figure 2d, feature cutting shows the better stability index than data cutting by statistical significance in every case ($\textit{p}<0.05$).

\subsubsection{Real Datasets}
\begin{figure}[htb!]
    \centering
    \includegraphics[width=0.9\columnwidth]{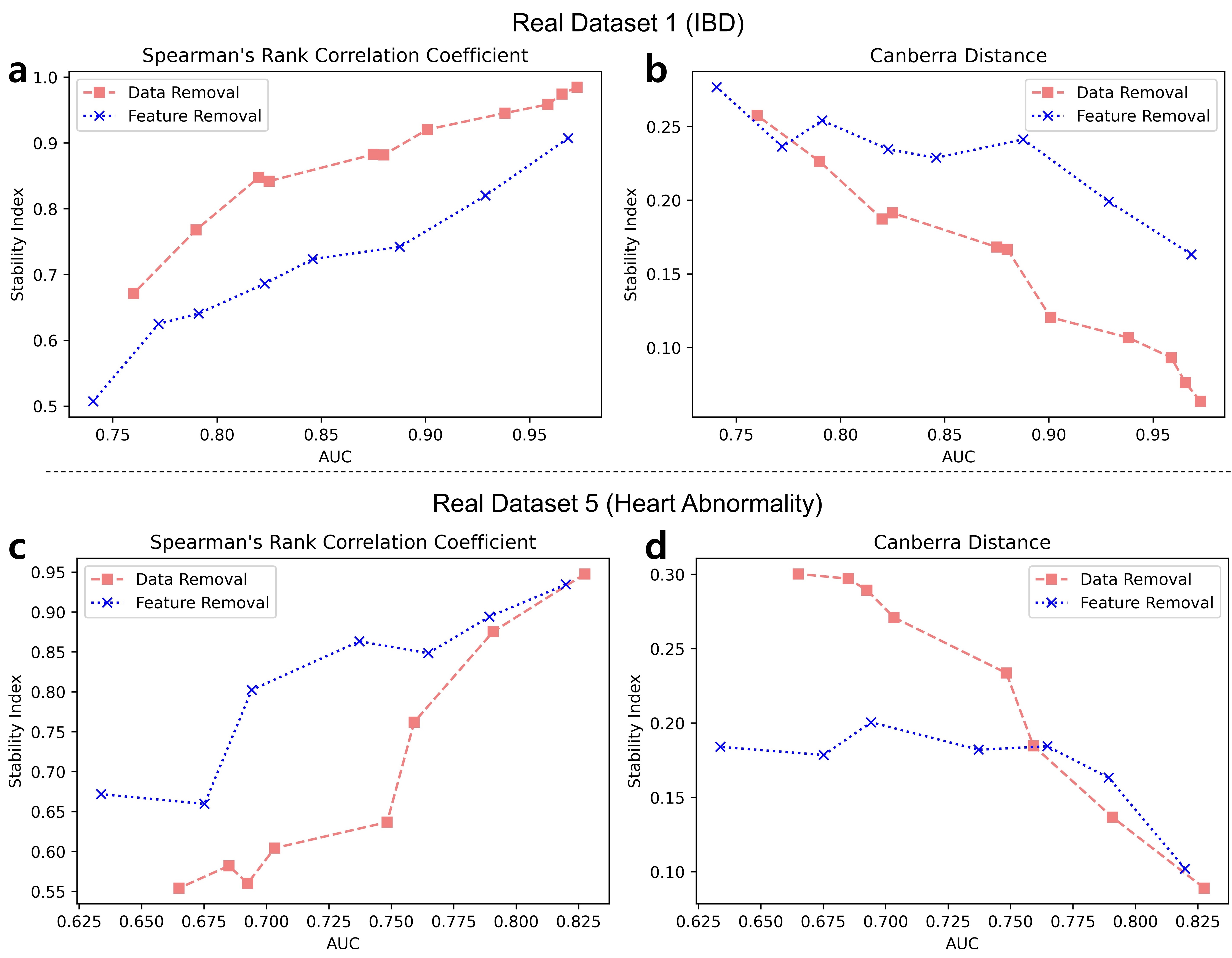}
    \caption{\textbf{Comparison of data cutting and feature cutting regarding SRCC and CD in experiment on real dataset 1 and 5.} Other stability indexes also show statistical significance, illustrated in Appendix 7 x-axis refers to AUC value and y axis to stability index.}
\end{figure}

To determine if similar patterns occur in real datasets, we performed the same analysis on six real datasets. Unlike synthetic datasets, real datasets have interdependent features with complex interactions, posing a disadvantage for feature cutting since the true feature rank may change as the set of features change. Thus, a conservative approach is used to assess whether feature cutting maintains better or similar stability compared to data cutting.

In all experiments, feature ranks deviate further from the original rank as performance decreases. The comparison between each degradation algorithm for each dataset is summarized in Appendix Table 3, based on the feature importance testing by linear regression model as in Figure 1d. Consistent patterns are observed between different stability indexes. In real datasets 3, 5, and 6, feature cutting shows better stability than data cutting, while the opposite is true for datasets 1, 2, and 4. However, all results lack statistical significance except for datasets 1 and 5.

In real dataset 1 (Figure 3a,b), data cutting generally shows better stability, though this difference diminishes as AUC decreases. In dataset 5 (Figure 3c,d), feature cutting shows evidently better stability. Figures of other datasets are available at Appendix 7. Overall, feature cutting generally shows similar or better stability compared to data cutting, except in dataset 1 (IBD). Feature cutting is more advantageous in terms of stability compared to data cutting as the comparison is done in smaller AUC ranges.

\subsection{Effect of Correlated Features}

\begin{figure}[htb!]
    \centering
    \includegraphics[width=0.9\columnwidth]{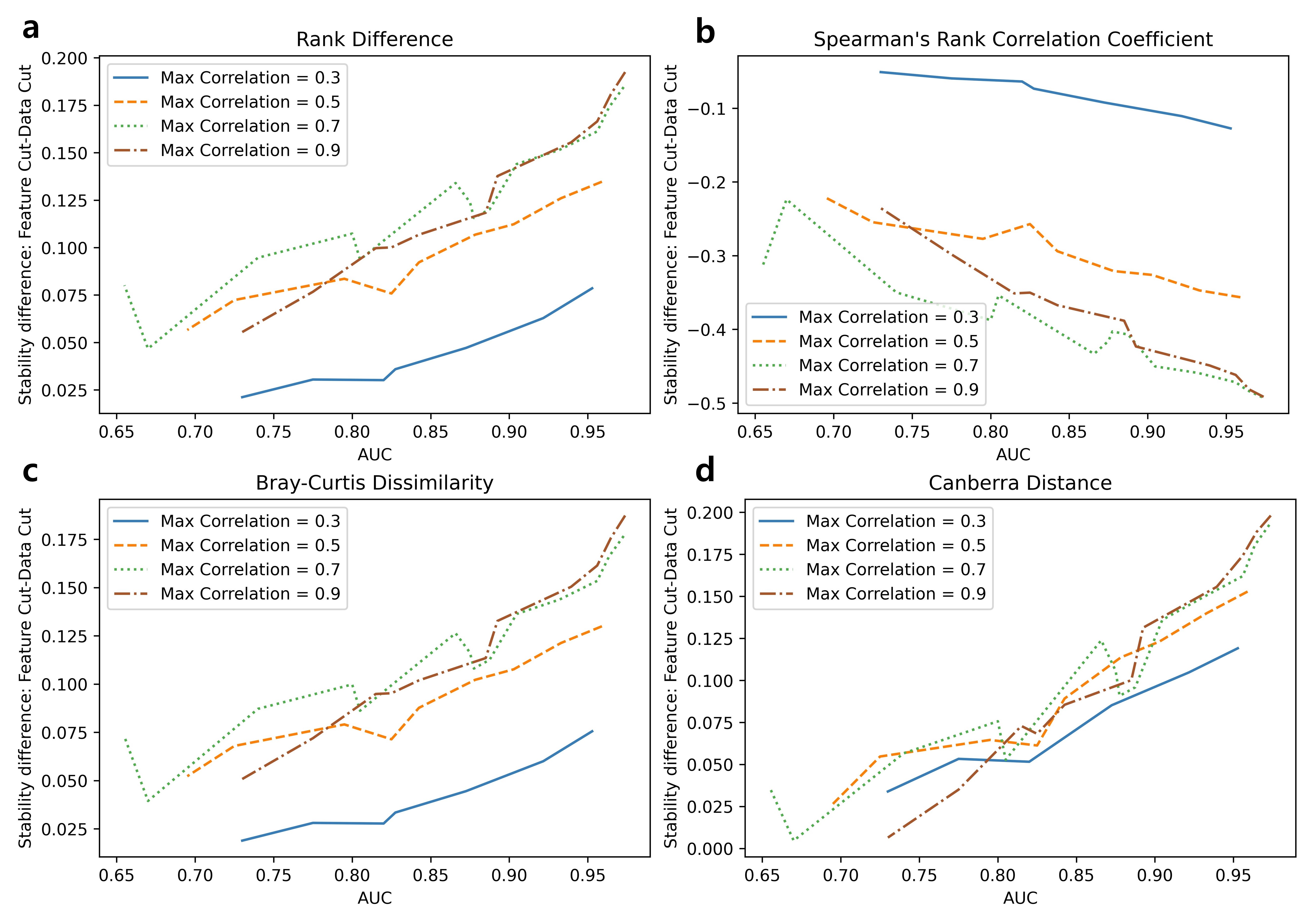}
    \caption{\textbf{Stability differences between feature cutting from data cutting after deleting correlated features higher than max correlation.} x-axis represents performance value (AUC), and y-axis represents stability index from feature cutting subtracted with the same index from data cutting. Each color represents maximum correlation among the feature group.}
\end{figure}
Since feature cutting does not show evident superiority over data cutting in some real datasets, further experimentation is needed to address this disadvantage. Although interactions cannot be directly calculated, correlations can partially represent interaction levels. To ensure consistent comparisons, highly correlated features are cut in stages, allowing subsequent experiments to focus on more independent features.

We repeated the performance degradation experiment of removing correlations between features. When two features exceeded the correlation cut standard, the weaker feature was deleted to minimize interactions. Results are shown in Appendix Table 4. To understand the impact of the correlation cut standard, a univariate regression analysis was conducted, as described in Figure 2e. Dataset 3 was excluded due to the already few correlations between features.

The impact of removing correlations in dataset 1 is illustrated in Figure 4. To compare stability indices at different AUCs, the stability index is interpolated for each degradation algorithm, and differences are measured in overlapping performance ranges. The difference between algorithms decreases as more correlations are removed. In other real world datasets, as more correlations are excluded, feature cutting generally exhibits better stability compared to data cutting, as shown in Appendix Table 4. Among the 20 correlation coefficients (5 datasets $\times$ 4 stability indexes), 16 indicate that removing more correlations benefits feature cutting. In the four opposite cases, the same pattern is observed, except that stability changes abruptly only when the correlation cut-off standard is 0.3, leaving too few features.

\subsection{Analysis of Feature Importance Distribution }

\begin{figure}
    \centering
    \includegraphics[width=0.95\columnwidth]{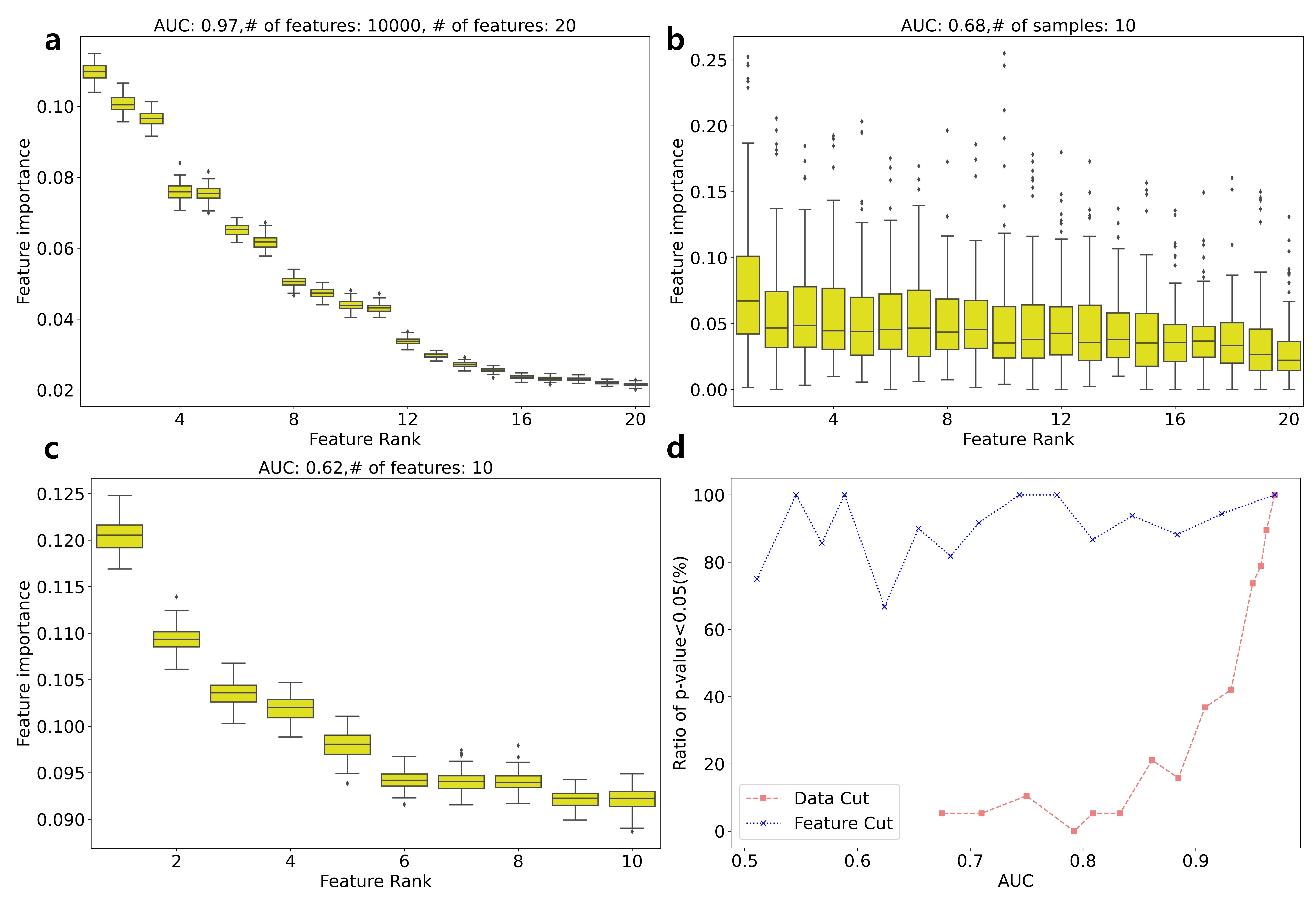}
    \caption{\textbf{Feature importance distribution of generated dataset 1.} In box plot (a, b, c), X-axis represents feature rank; y-axis represents value of feature importance (Gini impurity). Black dots show outliers from box plot. Feature importance of each feature for a) entire data samples and features, b) only with 10 samples (AUC=0.68), c) only with 10 features (AUC=0.62). d) Ratio of relationships where the feature importance values of adjacent features are distinguished by statistical significance ($\textit{p}<0.05$).}
\end{figure}

From Sections 3.1 and 3.2, we conclude that data cutting is the primary reason why performance degradation harms the validity of feature importance. To explore this, we analyzed the feature importance values, not just the ranks, to understand differences between degradation algorithms, as illustrated in Figure 5. The box plot shows Gini-impurity values for each feature across 100 bootstraps, with features aligned by their rank in each experiment.

With the full dataset (Figure 5a), feature importance distributions have few outliers and align with the true feature rank. With data cutting (Figure 5b), the distributions show many outliers, high variability, and no clear superiority among ranks. In contrast, feature cutting (Figure 5c) results in smaller variability and fewer outliers compared to data cutting.

To statistically validate if feature cutting distinguishes feature importance better than data cutting at similar performance levels, we compared feature importance values between adjacent ranks using a paired t-test (if both groups satisfy normality by the Shapiro-Wilk test) or the Wilcoxon rank-sum test, as illustrated in Figure 2f. In feature cutting (Figure 5d), the importance between features is clearly distinguished, regardless of performance, with most cases being statistically significant. However, in data cutting, machine learning fails to distinguish feature roles as performance decreases. The same analysis for real datasets is shown in Appendix 9. In every area and dataset, feature cutting shows a higher ratio of statistical significance.

\section{Probabilistic Approach}

To infer the theoretical mechanism behind the results, we calculate the theoretical probability that a training dataset has essential samples to differentiate the importance of features. We assume the population with \(M\) data points and \(N\) features where \(M\) and \(N\) are very large. We assume the same model used to generate synthetic data with independent features, except that the features are binary:
\[
y_m = \sum_{n=1}^{N} a_n x_{nm},\quad  x_{nm} \in \{0, 1\} \quad  \text{(Eq. 1)}
\]
where \(x_{nm}\) denotes the \(n\)th feature value of the \(m\)th sample, \(a_n\) is the feature importance value for a given feature, and if \(y_m > \mu\) then \(f(y_m) = 1\), else \(f(y_m) = 0\) with the threshold \(\mu\) to choose the class label. Feature coefficients are assumed to be proportional to \(\{1, 2, \ldots, N-1, N\}\) scaled to satisfy \(\sum_{n=1}^{N} a_n = 1\)  \text{(Eq. 2)}.

Let \(k\) be the number of samples randomly selected as a training dataset from the population (\(k \ll M\)). We want to calculate the probability \(p\) that the training dataset includes information to differentiate the relative superiority of feature \(i\)th with respect to feature \(j\)th without loss of generality (\(a_i < a_j\)). 

To distinguish the \(i\)th feature and \(j\)th feature, the model needs samples in which the importance gap between the \(i\)th feature and \(j\)th feature has a decisive impact on deciding labels. This happens when there are at least 2 samples among the \(k\) in the training set for which the \(i\)th feature and \(j\)th feature values are \((1,0)\) and \((0,1)\) respectively, and for which the following holds:
\[
\mu - a_j < \sum_{n=1}^{N} a_n x_{nm} - a_i x_{im} - a_j x_{jm} < \mu - a_i \quad \text{(Eq. 3)}
\]

In Appendix 10, we demonstrate that if \(N\) is sufficiently high and features \(x_n\) follow the Bernoulli distribution, then \(y_m\) is a realization of a variable \(Y\) that can be approximated by a Gaussian distribution with mean 0.5 and variance \(\frac{1}{3N}\). Therefore, the probability \(p\) that \(Y\) is in the range \((\mu - a_j, \mu - a_i)\) is proportional to \(a_j - a_i\) though it may not be linearly proportional.

Therefore, we can assume that the number of samples within the entire population satisfying Equation 3 is equal to \(M \cdot \left( \frac{\mu - a_i - \mu - a_j}{1 - a_i - a_j} \right) \approx M \cdot (a_j - a_i)\). Note that the denominator is approximated to 1 as the \(a_i + a_j\), the sum of the feature importance of two features, is assumed to be relatively very small compared to the entire sum of feature importance 1.

To confirm the superiority of the \(j\)th feature over the \(i\)th feature, there should be at least two samples from \(M \cdot (a_j - a_i)\) in which the \(i\)th feature and \(j\)th feature values are \((1,0)\) and \((0,1)\). Supposing we pick \(d > 1\) number of samples from the entire population of \(M\) samples, the probability \(g_d\) in which both combinations are equal to with some combinatorial calculus (Appendix 11):

\[
g_d = 1 - \left(\left( \frac{3}{4} \right)^d+\left( \frac{3}{4} \right)^d-\left( \frac{1}{2} \right)^d \right) \quad  \text{(Eq. 4)}
\]

To calculate \(p\), we can consider how many combinations include the information to distinguish the two features among the entire set of combinations selecting \(k\) samples from \(M\):

\[
p = \frac{\sum_{d=2}^{\min(M \cdot (a_j - a_i), k)} g_d {M - M \cdot (a_j - a_i)\choose k-d}{M \cdot (a_j - a_i)\choose d}}{\binom{M}{k}} \quad \text{(Eq. 5)}
\]

As we are picking \(d\) number of samples from the samples fulfilling Eq. 3, \(d\) can be at least 2 to \(k\) (if \(k\) is bigger than the number of samples fulfilling the condition, it should be no bigger than that). The term $\binom{M \cdot (a_j - a_i)}{d}$ represents the combinations where picking \(d\) number of samples from the samples which fulfill Equation 3. As \(d\) number of samples are picked from this group, the other \(k - d\) number of samples will be picked from the group where it does not satisfy Eq. 3, resulting in the term $\binom{M - M \cdot (a_j - a_i)}{k-d}$. Because we can suppose a high value for the entire dataset size \(M\), we can assume \(M \cdot (a_j - a_i)\) to be bigger than \(k\). Thus,



{
\footnotesize
\begin{align*}
p=
& \frac{\sum_{d=2}^{k} g_d \binom{M - M \cdot (a_j - a_i)}{k-d} \binom{M \cdot (a_j - a_i)}{d}}{\binom{M}{k}} 
\tag{Eq. 6}
\end{align*}
}

\begin{figure*}
    \centering
    \includegraphics[width=14cm]{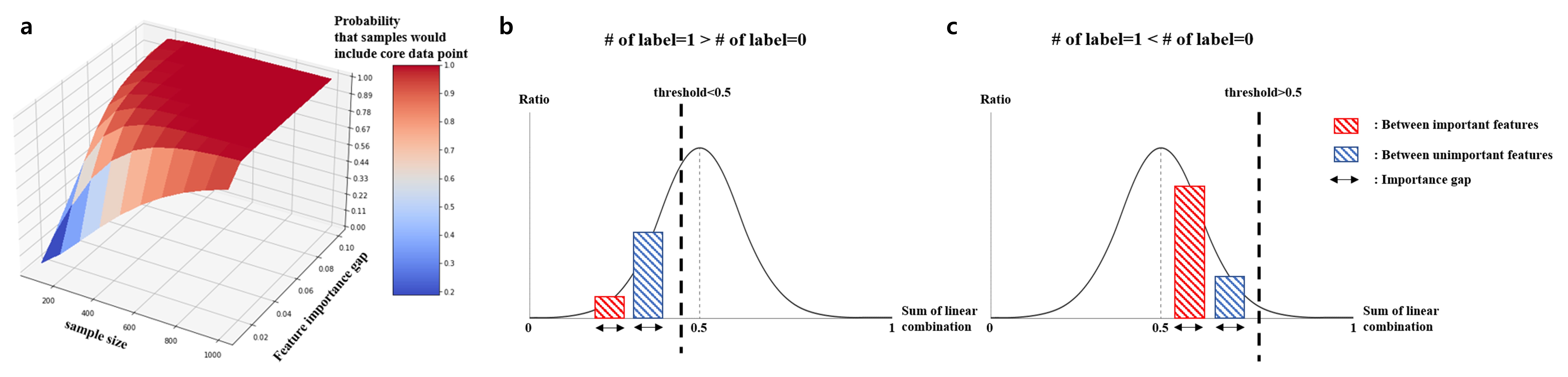}
    \caption{\textbf{Theoretical analysis in data-point view.} a) 3D plot to illustrate probability for sampled dataset to contain essential samples that enable distinction between two features.  control sample size (x-axis), and feature importance gap \((a_j - a_i)\) (y-axis) b), c) Probability distribution for linear combination sum to satisfy the condition 1}
\end{figure*}

To observe the pattern from the equation, we run a simulation with \(M = 1000\), \(N = 20\). Figure 6a illustrates the distribution of probability \(p\) that the training set of \(k\) examples include core data points. As the sample size decreases, the probability of including essential information in a sample group also decreases. Consequently, smaller sample sizes are more likely to miss core information needed to differentiate between features, leading to outliers. Since samples are selected randomly, outliers occur in a probabilistic way.

Lack of features can be modeled, supposing we only know a limited number of features. However, the number of features selected from the population will be not included in the final equation, \(p\). Although the performance may drop, the ratio of samples which satisfy the Eq.3 sustains.

\(p\) is theoretically at its maximum value because there is a premise behind this section that either 1) the feature importance values of other features are well known to the model enabling it to focus the usage of the information that distinguishes the \(i\)-th feature and \(j\)-th feature or 2) there are at least two samples satisfying the Eq.3 but with all the same feature values except for \(i\) and \(j\). However, this premise is influenced by whether we have enough data to determine the importance of other variables and by the extent to which the population can inherently be represented by a certain number of features, regardless of how many variables we selected.

\section{Discussion and Conclusion}
Countless studies have established that insufficient data is detrimental to machine learning \cite{rajput2023evaluation, balki2019sample, cui2018effect}. However, this leads to the incorrect expectation that performance is directly proportional to the validity of feature importance, limiting the use of feature importance in low-performing machine learning model, but allowing only statistical tests which cannot consider non-linear relationship. Our experiments with generated datasets and theoretical analysis demonstrate that low performance due to a lack of features does not harm validity and still provides valid relative comparisons between features. In real dataset experiments, we cannot rigorously prove that feature cutting has no impact on the validity of feature importance. However, by reducing the interactions, we achieve similar results in real datasets, proving that low performance due to a lack of features does not harm validity as much as a lack of data. Therefore, feature importance analysis should be considered independently and can be analyzed alongside statistics. Instead of focusing solely on performance, two approaches can be considered to validate the use of interpretable AI:

1) Assessing data sufficiency: Machine learning performance may plateau after a certain sample size \cite{rajput2023evaluation, faber2014sample}. Rajput et al. suggested evaluating sample size using effect sizes and performance. Alternatively, sample sufficiency can be assessed by proportionally cutting data size and observing performance changes. If performance remains stable with small cuts but drops significantly after larger cuts, low performance is likely due to a lack of features rather than data.

2) Analyzing Feature Importance Values: In Section 3.3, we found that feature importance values have many outliers and adjacent features are not well distinguished in data cutting. Instead of ranking features, applying statistical tests to see whether important features have significantly higher importance than others will provide valid results even in low-performing models \cite{lee2023suggestion, lee2024cleshcomprehensiveliteralexplanation}.

\subsection{Is it easier to distinguish between important features or between unimportant features?}


As explained in the Section 4 and Appendix 10, the probability that a sample satisfies the Eq.3 follows a normal distribution with a mean value of 0.5 and a variance that is inversely proportional to the entire feature number. Figure 7 illustrates the probability for selecting a sample that satisfies the given range. The probability can be calculated via integrating the density probability for the range $(\mu - a_j, \mu - a_i)$. If the threshold $\mu$ is smaller than the mean value 0.5, the integrated value is higher if $a_j$ is small which indicates that the probability will be higher when two unimportant features are compared. In other cases where $\mu$ is bigger than 0.5, the probability will be higher when two important features are compared unless $a_j$ is very huge such that $\mu - a_j$ is smaller than 0.5.

Real datasets often have more negative than positive samples. For medical datasets, negative samples (zeros) are easier to obtain than positive samples (ones), where one is the prediction target and zero is the normal case. This high threshold value makes important features easier to compare than unimportant ones. In our experiment, generated dataset 2 has more negative labels, while generated dataset 3 has more positive labels. Appendix 12 compares the feature importance distribution of each dataset, showing that the importance of the top 5 features is higher in generated dataset 2 but smaller after the 5th rank.

Class imbalance is well known to introduce bias in machine learning classification \cite{wu2023rethinking,ferrari2021addressing}, but not many studies have focused on bias in feature importance. Although the analysis in this section is derived exclusively from a data sampling standpoint, it provides important insights on this matter.

\subsection{Why do weak regression predictors tend to overestimate the importance of weak features?}

A `bias phenomenon' has been reported where linear classification models using gradient descent tend to overestimate the importance of weakly to moderately predictive features when training samples are insufficient \cite{leino2018feature}. In section 3.3, we observe that the importance distribution becomes flatter when performance is degraded by data cutting. This occurs in both simulation and real datasets using random forest, a tree-based ensemble model. Section 4 details the probability calculations that quantify the likelihood of this phenomenon from a sampling perspective. Considering the bias phenomenon, we see that when data size is insufficient, feature importance validity is compromised: the rank between features can be misunderstood, and the gap between important and unimportant features can be underestimated.

\bibliographystyle{plainnat}
\bibliography{article}

\end{document}